\documentclass[11pt,a4paper]{article}
\usepackage[hyperref]{acl2021}
\usepackage{times}
\usepackage{latexsym}
\usepackage{graphicx}
\usepackage{amsmath}
\usepackage{booktabs}

\usepackage{microtype}

\aclfinalcopy % Uncomment this line for the final submission
\def\aclpaperid{3482} %  Enter the acl Paper ID here

\title{DynaEval: Unifying Turn and Dialogue Level Evaluation}  

\author{Chen Zhang$^{\dag,\star}$ \quad Yiming Chen$^{\dag}$ \quad Luis Fernando D’Haro$^\ddag$ \quad Yan Zhang$^\dag$ \\ \quad \textbf{Thomas Friedrichs}$^\star$ \quad \textbf{Grandee Lee}$^\dag$ \quad \textbf{Haizhou Li}$^{\dag, \star\star}$     \\
  $^\dag$National University of Singapore \quad $^\star$Robert Bosch (SEA), Singapore \\
  $^\ddag$Universidad Politécnica de Madrid, Spain \quad $^{\star\star}$Kriston AI Lab, China \\
        \tt \{chen\_zhang,yiming.chen,grandee.lee\}@u.nus.edu, \\
        \tt \{haizhou.li,eleyanz\}@nus.edu.sg, \\
        \tt luisfernando.dharo@upm.es, thomas.friedrichs@sg.bosch.com
  }
\date{}

\begin{document}
\maketitle
\begin{abstract}
A dialogue is essentially a multi-turn interaction among interlocutors. Effective evaluation metrics should reflect the dynamics of such interaction. Existing automatic metrics are focused very much on the turn-level quality, while ignoring such dynamics. To this end, we propose DynaEval\footnote{\url{https://github.com/e0397123/DynaEval}}, a unified automatic evaluation framework which is not only capable of performing turn-level evaluation, but also holistically considers the quality of the entire dialogue. In DynaEval, the graph convolutional network (GCN) is adopted to model a dialogue in totality, where the graph nodes denote each individual utterance and the edges represent the dependency between pairs of utterances. A contrastive loss is then applied to distinguish well-formed dialogues from carefully constructed negative samples. Experiments show that DynaEval significantly outperforms the state-of-the-art dialogue coherence model, and correlates strongly with human judgements across multiple dialogue evaluation aspects at both turn and dialogue level. 
\end{abstract}

% \section{Credits}

% This document has been adapted by Roberto Navigli
% from the instructions for earlier ACL, NAACL and EMNLP proceedings, including those for 
% EMNLP 2020 by Yulan He,
% ACL 2020 by Steven Bethard, Ryan Cotterrell and Rui Yan, 
% ACL 2019 by Douwe Kiela and Ivan Vuli\'{c},
% NAACL 2019 by Stephanie Lukin and Alla Roskovskaya, 
% ACL 2018 by Shay Cohen, Kevin Gimpel, and Wei Lu, 
% NAACL 2018 by Margaret Michell and Stephanie Lukin,
% 2017/2018 (NA)ACL bibtex suggestions from Jason Eisner,
% ACL 2017 by Dan Gildea and Min-Yen Kan, 
% NAACL 2017 by Margaret Mitchell, 
% ACL 2012 by Maggie Li and Michael White, 
% ACL 2010 by Jing-Shing Chang and Philipp Koehn, 
% ACL 2008 by Johanna D. Moore, Simone Teufel, James Allan, and Sadaoki Furui, 
% ACL 2005 by Hwee Tou Ng and Kemal Oflazer, 
% ACL 2002 by Eugene Charniak and Dekang Lin, 
% and earlier ACL and EACL formats written by several people, including
% John Chen, Henry S. Thompson and Donald Walker.
% Additional elements were taken from the formatting instructions of the \emph{International Joint Conference on Artificial Intelligence} and the \emph{Conference on Computer Vision and Pattern Recognition}.

\section{Introduction}

\label{sec:introduction}
Modern dialogue systems~\citep{smith-etal-2020-put,zhang-etal-2020-dialogpt,adiwardana2020towards} leveraging large-scale language model pre-training~\citep{devlin-etal-2019-bert,radford2019language} are capable of generating fluent and contextually relevant utterances. Yet, they still face difficulties in mimicking human conversations in the sense that they lack certain conversation-level attributes, such as coherence~\citep{cervone2018coherence}, consistency~\citep{welleck-etal-2019-dialogue,nie2020like}, diversity~\citep{li-etal-2016-diversity,wu-etal-2020-diverse} and engagement~\citep{ghandeharioun2019approximating,ghazarian2020predictive}. One of the main reasons is the dearth of effective dialogue-level evaluation mechanisms to guide the studies and to monitor progress.  

Commonly used static metrics, such as BLEU~\citep{papineni-etal-2002-bleu}, METEOR~\citep{denkowski-lavie-2014-meteor} and ROUGE~\citep{lin-2004-rouge}, correlate poorly with human judgements~\citep{liu-etal-2016-evaluate} rendering them unsuitable for dialogue evaluation. 
While some recent automatic dialogue evaluation metrics~\citep{ghazarian-etal-2019-better,mehri-eskenazi-2020-usr,huang-etal-2020-grade,zhang-etal-2021-dscore} demonstrate strong correlations with human judgement at the turn-level,  they only focus on context-response pairs without explicitly modeling the interaction over an entire dialogue. To perform dialogue-level evaluation, we need to rely on the aggregation of turn-level scores over the dialogue as a proxy for a dialogue-level score.

Furthermore, a recent study by \citet{mehri-eskenazi-2020-unsupervised} found out that even though state-of-the-art chatbots outperform humans across multiple turn-level evaluation criteria, such as interestingness, engagement and specificity, their dialogue-level ratings like coherence, Likability and diversity are still far below human level. This further reinforces the idea that turn-level quality evaluation may be insufficient to assess the performance of open-domain dialogue systems. 

In this work, we address the problem of automatic open-domain dialogue evaluation by focusing on the quality of an entire dialogue. This is a departure from the way we frame the problem as a weakly supervised next sentence prediction~\citep{mehri-eskenazi-2020-usr,sato-etal-2020-evaluating} or language modeling tasks~\citep{nedelchev-etal-2020-language,pang-etal-2020-towards} for context-response pairs. To this end, we need to answer two important questions: (1) How to effectively represent the entire dialogue? (2) How to incorporate this dialogue-level knowledge into our evaluation framework? We propose DynaEval to provide meaningful dialogue-level representation with explicit modeling of the interactive dynamics among interlocutors, for a unified turn and dialogue level quality assessment.

The main contributions of this work include: (1) The unified turn and dialogue level evaluation represents a departure from turn-level evaluation scheme; (2) DynaEval is one of the first few metrics where dialogue level dynamics is considered with structured graph representation. (3) Empirical results show that DynaEval outperforms the state-of-the-art dialogue coherence model and strongly correlates with human judgements at both turn and dialogue level. 

\section{Related Work}
\label{sec:related_work}

\subsection{Open-ended Dialogue Evaluation}

\textbf{Turn-Level Evaluation }  
The current trend for automatic dialogue evaluation is shifting towards the reference-free paradigm. Lately, the research community has witnessed a surge in the automatic metrics along these lines. 
Many of them focus on evaluating naturalness of generated responses. Typical examples include perplexity~\citep{adiwardana2020towards}, USR-MLM~\citep{mehri-eskenazi-2020-usr} and GPT-2~\citep{radford2019language} based fluency metrics~\citep{nedelchev-etal-2020-language,pang-etal-2020-towards}. 

Another group of metrics evaluates contextual relevance of the responses. For example, RUBER~\citep{tao2018ruber}, BERT-RUBER\citep{ghazarian-etal-2019-better} and USR-DR~\citep{mehri-eskenazi-2020-usr} predict the relatedness between generated responses w.r.t the corresponding context by training a discriminative network to distinguish the original response from negative samples bootstrapped from the training set.~\citet{sato-etal-2020-evaluating} and~\citet{lan2020pone} provide a better sampling strategy for bootstrapping negative samples. 

Besides these two major aspects, there are many metrics for other qualities, such as adequacy~\citep{d2019automatic,zhang2021deep}, consistency~\citep{welleck-etal-2019-dialogue,dziri-etal-2019-evaluating-coherence}, engagement~\citep{ghazarian2020predictive}. 

Even though all these automatic metrics demonstrate strong correlation with human judgements, they are laser-focused on one aspect of the evaluation. In addition, they do not explicitly model the speaker-level and utterance-level interactions, which we believe is essential for the dialogue-level representation, and eventually benefits the dialogue evaluation task.  

\textbf{Interactive Evaluation } 
A popular human evaluation method is the interactive evaluation whereby human judges converse with dialogue systems and make the assessment at the end of the conversations~\citep{see-etal-2019-makes,finch-choi-2020-towards,li2019acute,deriu-etal-2020-spot}. It has been shown to be more reliable than turn-level static evaluation~\citep{mehri-eskenazi-2020-unsupervised}. 

There are few studies on fully automating this process.~\citet{ghandeharioun2019approximating} propose a self-play scenario where the dialog system chats with itself and a combination of three metrics measuring sentiment, semantic coherence and engagement respectively along the conversation trajectory is computed to approximate dialogue-level quality estimation.~\citet{mehri-eskenazi-2020-unsupervised} propose the FED metric, which evaluates the quality of a system utterance in an interactive setting by computing the likelihood of a particular follow-up utterance responded by dialoGPT~\citep{zhang-etal-2020-dialogpt}. Moreover,~\citet{sinha-etal-2020-learning} come up with MaUde, a reference-free metric tailored for online dialogue evaluation, which leverages a pre-trained DistilBERT~\cite{sanh2019distilbert} model to extract the semantic representation of dialogue turns and uses bidirectional LSTM to explicitly model the discourse structure.

While the interactive evaluation is more reliable than the turn-level static evaluation, it still relies on the aggregation of turn-level scores. An ideal approximation of the human evaluation process is a top-down approach whereby we examine the quality of the entire dialogue at macro level before zooming into the dialogue turns. Hence, a unified framework, which holistically models the entire dialogue, is highly sought after. 

\subsection{Dialogue Coherence}

Examining a dialogue at macro level is related to discourse coherence~\citep{halliday2014cohesion,grosz1995centering,barzilay2008modeling}, which considers whether a piece of text is in a consistent and logical manner, as opposed to a random collection of sentences. Dialogue is a special kind of discourse structure, of which coherence assessment is an essential part of quality evaluation.

Many studies have followed the standard discourse coherence evaluation protocol~\citep{cervone-riccardi-2020-dialogue,zhou2019hierarchical,mesgar-etal-2020-dialogue}. Very few have considered customizing their dialogue coherence models for evaluating the performance of dialogue systems. It is common to leverage supervised approaches~\citep{higashinaka2014evaluating,gandhe2016semi,cervone2018coherence,yi-etal-2019-towards}, that is closely linked to modeling with entities and dialogue acts~\citep{cervone-riccardi-2020-dialogue,zhou2019hierarchical,mesgar-etal-2020-dialogue}. 

Hence, we are motivated to study the application of dialogue coherence modeling for automatic dialogue evaluation by designing a self-supervised framework, without dependence on any human annotations for coherence features.  

\subsection{Graph Modeling of Dialogue}

Recently, the graph neural network (GNN) \citep{scarselli2008graph,thomas2018graph,schlichtkrull2018modeling} has been successfully applied in various dialogue applications. For example, ~\citet{ghosal-etal-2019-dialoguegcn} adopts GCN for utterance-level emotion recognition.~\citet{chen-etal-2018-structured} modeled structured dialogue policy with GNN and~\cite{qin2020co} proposes a joint framework leveraging graph attention network~\citep{velivckovic2017graph} for both dialogue act recognition and sentiment classification. 

GNN is useful for dialogue modeling, because the relative position of target and context utterances decides how past utterances influence future utterances and vice versa~\citep{ghosal-etal-2019-dialoguegcn}. The interaction of utterances can be effectively captured with a graph structure as long as they are connected by relation-aware edges. However, GNN has not been well studied for dialogue evaluation.  \citet{huang-etal-2020-grade} recently proposes the GRADE metric, leveraging graph modeling for turn-level coherence evaluation. The way we use GNN is different from \citet{huang-etal-2020-grade} because GRADE is focused on turn-level coherence evaluation while we are interested in a turn-dialogue joint evaluation. Furthermore, GRADE considers the keywords in context-response pairs, and we explicitly use graph structure to model the speaker and utterance level interaction within a dialogue.

\section{DynaEval Framework}
\label{sec:dynaeval}

DyanEval represents an integration of several ideas. It takes advantage of the structured graph representation of dialogues, useful information on the utterance and speaker level interaction. It is motivated by dialogue coherence modeling. 

\begin{figure*}[t]
    \centering
    \includegraphics[width=\linewidth]{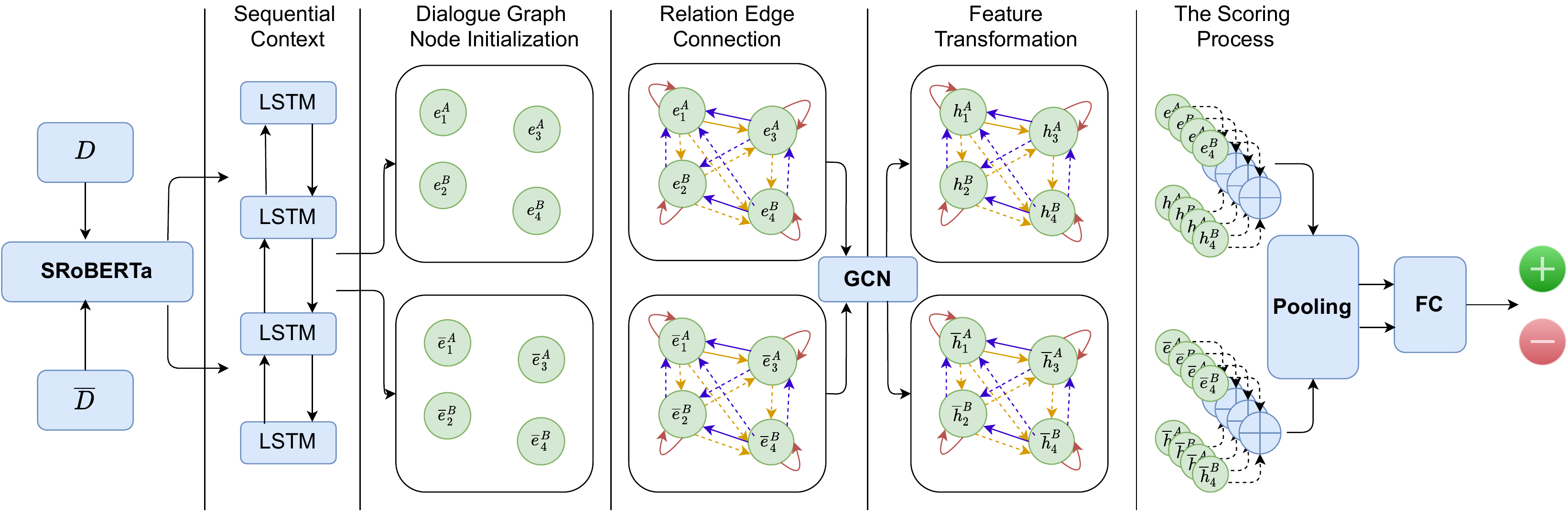}
    \caption{The architecture of DynaEval. The input is a pair of contrasting dialogues, $D$ and $\bar{D}$. The output is a unified score indicating whether $D$ is preferred than $\bar{D}$. Utterance-level representation derived from SRoBERTa model is used for dialogue graph node initialization. Different types of arrows in relation edge connection represent different types of relations: (1) Solid line denotes intra-speaker dependency. (2) Dotted line denotes inter-speaker dependency. (3) Red color means self-connection. (4) Purple color means connection from future utterances to previous utterances. (5) Yellow color means connection from previous utterances to future utterances. Since there are two speakers, A and B. Hence, there will be a total of 2 $\times$ 2 $\times$ 2 + 1 = 9 distinct relation types.} 
    \label{fig:dynaeval-architecture}
\end{figure*}

In this paper, we only consider dyadic dialogues, but the formulation can be easily generalized to multi-party conversations. Formally, let $A$ and $B$ denote the two speakers participating in the dialogue. A dialogue, $D$, consists of a sequence of $n$ utterances, $[u^{A}_{1}, u^{B}_{2}, \ldots, u^{A}_{n-1}, u^{B}_{n}]$\footnote{n is assumed to be even to simplify the mathematical expressions.}. Let $\bar{D}$ represent the negative dialogue sample obtained via various sampling strategies described in Section~\ref{subsec:training}. 

Figure~\ref{fig:dynaeval-architecture} illustrates the learning process of DynaEval in four steps\footnote{Note that all the operations from Section~\ref{subsec:dur} through Section~\ref{subsec:tsp} are illustrated with $D$. They are applied in the same way on $\bar{D}$.}: (1) Deriving contextualized representation, $\textbf{e}_i$, for utterances within $D$. (Section~\ref{subsec:dur}). (2) Constructing the directed dialogue graph. The nodes are initialized with $\textbf{e}_{i}$ and the edges between node pairs represent the speaker and temporal dependencies (Section~\ref{subsec:dgc}). (3) Generating utterance-level graph representation, $\textbf{h}_{i}$, via feature transformation to aggregate useful contextual information from all connected neighbours to the current node (Section~\ref{subsec:ft}). (4) producing a dialogue-level score, which indicates whether $D$ is preferred over $\bar{D}$ (Section~\ref{subsec:tsp}).

\subsection{Dialogue Utterance Representation}
\label{subsec:dur}
A sentence-encoder is needed to map the individual utterances within $D$ onto the vector space. Firstly, we fine-tune a RoBERTa-base pre-trained language model~\citep{liu2019roberta} with training data of the target dialogue domain, because task-adaptive fine-tuning of the pre-trained language model on the target domain data benefits the final performance~\citep{gururangan-etal-2020-dont, lee-li-2020-modeling}. Next, the mean pooling operation is performed on the token embeddings within each utterance of $D$ to derive their respective utterance-level representations. Formally, let SRoBERTa denotes the sentence encoder and $u^{*}_i$ in $D$ is mapped into vector representations, $\textbf{u}_i\in{\mathcal{R}^{d}}$, whereby
\begin{equation}
    \textbf{u}_i = \text{SRoBERTa}(u^{*}_i)
    \label{eq:input_org} \\
\end{equation}
Note that $*$ can be either speaker $A$ or speaker $B$. Then, to capture a more fine-grained temporal dependency among the utterances, a bidirectional LSTM is adopted to model the sequential flow of information within $D$. The context-aware utterance representation, $\textbf{e}_i$ is then obtained via:

\begin{equation}
    \textbf{e}_i = \overleftrightarrow{\text{LSTM}}(\textbf{e}_{i(+,-)1}, \textbf{u}_i)
    \label{eq:input_lstm} \\
\end{equation}

\subsection{Dialogue Graph Construction}
\label{subsec:dgc}
$D$ is represented with a directed graph, $\mathcal{G} = (\mathcal{V}, \mathcal{E})$. $\mathcal{V}$ is the sets of graph nodes and $\mathcal{E}$ is the set of edges, which reflects the contextual dependencies among utterance pairs.

\paragraph{Graph Nodes}
Each graph node corresponds to an utterance within $D$. Hence, for a dialogue with n utterances, $\mathcal{V} = \{v_1, v_2,\ldots,v_{n-1}, v_{n}\}$. All the graph nodes are initialized with utterance-level contextualized embeddings: $v_i = \textbf{e}_i$. 

\paragraph{Edges}
For short conversations, $\mathcal{G}$ will be a fully-connected graph whereby all graph nodes are connected to each other, including self-connection. The intuition is that short conversations tend to focus on a single topic and thus, each utterance is contextually dependent on all the other utterances in the dialogue. For long conversations, there may be frequent topic shifts. Distant utterances within the same dialogue may not be contextually relevant to the current utterance. 
Sometimes, adding more context leads to diminishing performance gain or even negative impact~\citep{zhong-etal-2019-knowledge}. Therefore, a context window length, $M$, is set, which means that $v_i$ is only connected to $v_j\in{\{v_{i-M}, v_{i-M+1},\ldots,v_i,v_{i+1},\ldots,v_{i+M}\}}$\footnote{For simplicity purpose, we do not explicitly include the cases when $i <= M$ or $i + M$ is greater than the total number of utterances in a dialogue in the formula.}. Let $v_{ij}\in{\mathcal{E}}$ denote the edge from $v_j$ to $v_i$. Each edge is associated with an edge weight, $a_{ij}$, and a relation type, $\theta_{ij}$. They are illustrated as follows:  

\textbf{Edge Weights } The edge weight determines the relative importance of the neighbour nodes w.r.t the current node. A similarity based attention module is applied to determine the edge weights. For a graph node, $v_i$, the set of weights, $a_i$, w.r.t all its incoming edges, should sum up to 1. The attention weight is formulated in the following way:
\begin{equation}
    \label{eq:egdge-weight}
    \begin{split}
a_{i} & = \text{softmax}(\textbf{e}^T_iW_{e}[\textbf{e}_{i-M},\ldots,\textbf{e}_{i+M}]), \\
& \text{where} \sum^{i+M}_{j=i-M}a_{ij} = 1, W_e\in{\mathcal{R}^{d\times{d}}} 
\end{split}
\end{equation}
More importance is placed upon neighbouring utterances on the same topic. Little attention is paid to the irrelevant utterances.

\textbf{Edge Relations } Following~\citep{ghosal-etal-2019-dialoguegcn}, there are two aspects to take into account when defining the relation types. One aspect is to capture speaker dependencies. This is because we want to model the interaction between the interlocutors in a dialogue. The other aspect is to consider the temporal dependencies. This pertains to the relative position of an utterance w.r.t another. The explicit modeling of such dependency is important since the ordering of utterances within a dialogue is an essential feature for learning dialogue coherence. With these considerations, the total number of distinct types of relations\footnote{Since we are considering dyadic dialogues, there are only two speakers involved. The formulation can be generalized to multi-party dialogue.} will be 2 ($u_i^*$ occurs before or after $u_j^*$) $\times$ 2 (either $u_i^A$ or $u_i^B$) $\times$ 2 (either $u_j^A$ or $u_j^B$) plus the self-connection ($i=j$). This is depicted with different arrows connecting the graph nodes in Figure~\ref{fig:dynaeval-architecture}. We define this set of 9 relation types as $\Theta$ and $\theta_{ij}\in{\Theta}$.

\subsection{Feature Transformation}
\label{subsec:ft}
This section describes the process of transforming the initial node representation, $\textbf{e}_i$, into both a speaker and context aware vector representation, $\textbf{h}_i$, which captures the dynamics of interaction w.r.t $u_i^*$. Basically, the whole process is a two-stage graph convolution. 

The first stage aggregates information from neighbourhood nodes to the current node $v_i$ based on the relation-aware transformation motivated by~\citep{schlichtkrull2018modeling} whereby edges of different relation types are associated with different transformation matrix, $W_\theta^{'}$:
\begin{equation}
    \label{eq:trans-one}
    \begin{split}
        \textbf{h}_i^{'} & = \sigma(\sum_{\theta\in{\Theta}}\sum_{j\in{S^\theta_i}}\frac{a_{ij}}{c_{i,\theta}}W_\theta^{'}\textbf{e}_j + a_{ii}W_0^{'}\textbf{e}_i) \\
    & \text{for } i = 1,2,\ldots,n
    \end{split}
\end{equation}
In Equation~\ref{eq:trans-one}, $\textbf{h}_i^{'}$ is the intermediate node representation and $\sigma$ denotes the activation function, such as ReLU. $S^{\theta}_{i}$ represents the set of indices of nodes connected to $v_i$ with their edges $v_{ij}$ having the relation type $\theta\in{\Theta}$. $a_{ij}$ and $a_{ii}$ are the edge weights of $v_{ij}$ and $v_{ii}$ respectively. $W_\theta^{'}\in{\mathcal{R}^{d^{'}\times{d}}}$ and  $W_0^{'}\in{\mathcal{R}^{d^{'}\times{d}}}$ are learnable parameters of the feature transformation. $c_{i,\theta}$ is a problem specific normalization constant, which can be set as a learnable parameter or fixed in advance.

The second stage applies another graph convolution operation on the intermediate node representation, $\textbf{h}_i^{'}$ and the final node representation, $\textbf{h}_i$ is obtained via: 
\begin{equation}
    \label{eq:trans-two}
    \begin{split}
        \textbf{h}_i & = \sigma(\sum_{j\in{S^\theta_i}}W^{''}\textbf{h}_j^{'} + W_0^{''}\textbf{h}_i^{'}) \\
    & \text{for } i = 1,2,\ldots,n
    \end{split}
\end{equation}
where $W^{''}\in{\mathcal{R}^{d^{''}\times{d^{'}}}}$ and $W_0^{''}\in{\mathcal{R}^{d^{''}\times{d^{'}}}}$ are two learnable parameters in the second stage of feature transformation. 

Through Equation~\ref{eq:trans-one} and Equation~\ref{eq:trans-two}, relevant contextual information from neighbouring nodes is effectively accumulated to the current node while irrelevant information is filtered out.

\subsection{The Scoring Process}
\label{subsec:tsp}
In the scoring step, $\textbf{h}_i$ is first concatenated with $\textbf{e}_i$ to obtain the final utterance representation, $\textbf{g}_i$. Next, a mean pooling layer is applied on all the utterance representations in a conversation to derive the dialogue-level representation, $\textbf{o}$:
\begin{equation}
    \label{eq:pooling}
    \textbf{o} = \frac{\sum_{i=1}^{n}\textbf{g}_i}{|\sum_{j=1}^{n}\textbf{g}_j|}
\end{equation}
$\bar{\textbf{o}}$, which corresponds to $\bar{D}$, is obtained in the same way. A unified score, $s_{dial}$ or $s_{\bar{dial}}$, is derived by passing $\textbf{o}$ or $\bar{\textbf{o}}$ through a fully-connected layer.

\subsection{Training Setup}
\label{subsec:training}
\paragraph{Learning Objective}
Inspired by the preference learning approaches, the label, $y$ for the $D$ and $\bar{D}$ pair is defined as:
\begin{equation}
      y =
    \begin{cases}
      1 & \text{if $D$ is preferred over $\bar{D}$}\\
      -1 & \text{if $\bar{D}$ is preferred over $D$}
    \end{cases}  
\end{equation}
The margin ranking loss function is adopted to train DynaEval. 
\begin{equation}
    \mathcal{L} = \max(0, -y * (s_{dial} - s_{\bar{dial}}) + 1)
\end{equation}

\paragraph{Sampling Strategy}
Two negative sampling strategies are explored in this paper to construct $\bar{D}$: Utterance Replacement (UR) and Speaker Level Utterance Shuffling (SS).

\textbf{Utterance Replacement (UR)} 
An utterance randomly selected from a dialogue is replaced with another utterance randomly chosen from a different dialogue. This sampling strategy perturbs a dialogue at the semantic level. An utterance from a different dialogue is considered topically in-congruent w.r.t the current dialogue context. It breaks down the current dialogue by suddenly injecting irrelevant information.

\textbf{Speaker Level Utterance Shuffling (SS) } 
With this strategy, the order of utterances from one speaker in a dialogue is kept the same while that from another speaker is shuffled. SS changes the coherence structure of a dialogue w.r.t specific speaker. This strategy is motivated by~\citep{healey2014divergence}, which adopts a "Chance Other" method to measure how much syntactic and lexical repetition of a speaker happen by chance. The reason why we do not randomly permute the order of all utterances in the dialogue is because random permutation of all utterances is a very simple discrimination task.

\section{Experiments}
\label{sec:experiment}

In this work, we consider two experiment settings to assess the effectiveness of DynaEval. The first setting (Section~\ref{subsec:ddt}) is similar to the studies on dialogue coherence~\citep{cervone2018coherence,mesgar-etal-2020-dialogue} where accuracy score is applied to evaluate its discrimination capability in distinguishing original dialogues from negative samples. The second setting (Section~\ref{subsec:correlation-analysis}) is to evaluate its dialogue-level and turn-level judgement capability via correlation analysis on the human-chatbot conversational datasets. The domain of the evaluation set is different from that of human-human conversation datasets that DyanEval is trained on.

\subsection{Dialogue Datasets}
\label{subsec:dataset}
Three bench-marking open-domain dialogue datasets are included in our experiments, Empathetic Dialogue~\citep{rashkin-etal-2019-towards}, ConvAI2 PERSONACHAT~\citep{zhang-etal-2018-personalizing,dinan2019second} and DialyDialog~\citep{li-etal-2017-dailydialog}. For training, we remove dialogues containing less than 4 utterances or more than 30 utterances. Statistics of the three human-human dialogue corpora after filtering is presented in Table~\ref{tab:dataset-statistics}.

\textbf{Empathetic Dialogue} 
is designed for mimicking the real-life human conversation scenario whereby the interlocutors need to recognize and acknowledge the others' feelings in the conversation. This dataset pertains to the short conversation scenario where interlocutors stick to a single topic.

\textbf{ConvAI2 PERSONACHAT } 
is a crowd-sourced dataset where each pair of interlocutors try to get to know each other by conditioning their conversations on their respective persona profile provided in prior. The dataset contains more number of turns per dialogue as compared to Empathetic Dialogue. Hence, topic shift is more likely to occur within a dialogue and this simulates the long conversation scenario mentioned in Section~\ref{subsec:dgc}. 

\textbf{DailyDialog } 
is a high-quality human-human conversation dataset, which reflects our day-to-day communications and covers different topics about our daily life, such as relationship and health. The average dialogue length of DailyDialog lies in the middle of that of Empathetic Dialogue and ConvAI2. Topic shift in the conversations of DailyDialog occurs less frequently as compared to those in ConvAI2.

\begin{table}[!t]
\normalsize
\centering
\resizebox{\linewidth}{!}{
\begin{tabular}{l|ccc}
\toprule
\textbf{Empathetic Dialogue} & \multicolumn{1}{c}{\textbf{training}} & \multicolumn{1}{c}{\textbf{validation}} & \multicolumn{1}{c}{\textbf{test}} \\ \midrule
\#dialog & 19,531 & 2,768 & 2,547 \\
\#turn   & 84,160 & 12,075 & 10,973 \\
\#word   & 1,306,060 & 201,816 & 194,772 \\
\#avg turn per dialogue   & 4.31 & 4.36 & 4.31 \\ 
\#avg words per dialogue   & 66.87 & 72.91 & 76.47 \\ \midrule
\textbf{ConvAI2} & \multicolumn{1}{c}{\textbf{training}} & \multicolumn{1}{c}{\textbf{validation}} & \multicolumn{1}{c}{\textbf{test}} \\ \midrule
\#dialog & 17,878 & 1,000 & - \\
\#utterance   & 262,626 & 15,566 & - \\
\#word   & 3,068,672 & 189,374 & - \\
\#avg turn per dialogue  & 14.69 & 15.57 & - \\ 
\#avg words per dialogue   & 171.64 & 189.37 & - \\ \midrule
\textbf{DailyDialog} & \multicolumn{1}{c}{\textbf{training}} & \multicolumn{1}{c}{\textbf{validation}} & \multicolumn{1}{c}{\textbf{test}} \\ \midrule
\#dialog & 10,245 & 933 & 918 \\
\#utterance & 84,916 & 7,908 & 7,536 \\
\#word & 1,189,527  & 109,172 & 106,627 \\
\#avg turn per dialogue & 8.29 & 8.48 & 8.21 \\ 
\#avg words per dialogue & 116.11 & 117.01 & 116.15 \\ \bottomrule
\end{tabular}
}
\caption{Human-Human Dialogue Corpora Statistics}\label{tab:dataset-statistics}
\end{table}

\subsection{The Dialogue-level Discrimination Task}
\label{subsec:ddt}
Similar to the previous works~\citep{cervone-riccardi-2020-dialogue,mesgar-etal-2020-dialogue}, 20 perturbations are created for each dialogue w.r.t both UR and SS. For each perturbation, two pairs are formed, $\{D, \bar{D}\}$ with label $y = 1$ and $\{\bar{D}, D\}$ with label $y = -1$. Then, we train, fine-tune, and evaluate DynaEval on the training, validation, and test sets for each sampling strategy. Note that all these sets are constructed with the same perturbation method.

\begin{table*}[!ht]
\centering
\resizebox{\linewidth}{!}{
\begin{tabular}{@{}lcccccccc@{}}
\toprule
& \multicolumn{2}{c}{Empathetic}   &     & \multicolumn{2}{c}{ConvAI2} & &
\multicolumn{2}{c}{DailyDialog}
\\\cmidrule{2-3} \cmidrule{5-6} \cmidrule{8-9}
Model    & \multicolumn{1}{c}{UR} & SS    &       & \multicolumn{1}{c}{UR} & SS  & &
\multicolumn{1}{c}{UR} & SS \\ \midrule
RANDOM   & 50.07 & 50.07   &     & 50.25 & 50.25 &    & 50.17 & 49.62         \\
CoSim    & 63.54 &  63.33  &    & 68.79 & 92.93   &    & 69.59 & 63.80      \\
S-DiCoh  & 80.33 ± 2.83 & 86.04 ± 0.31 && 66.80 ± 1.93 & 90.35  ± 0.08 && 83.67 ± 0.41 & 84.92  ± 0.70 \\ \midrule
DynaEval & \textbf{94.30 ± 0.07} & \textbf{90.37 ± 0.37} && \textbf{85.23 ± 0.96}  & \textbf{98.65 ± 0.29} && 
\textbf{91.89 ± 0.58}  & \textbf{91.65 ± 0.62} \\ \bottomrule
\end{tabular}
}
\caption{The accuracy (\%) of DynaEval vs baselines on the test sets of Empathetic Dialogue and DailyDialog as well as the validation set of ConvAI2. UR \& SS are the sampling strategies defined in Section~\ref{subsec:training}. Experiments involving training are repeated five times with different random seeds for model weights initialization. The average and standard deviation are reported in the table.}
\label{tab:coh-results}
\end{table*}

\textbf{Baselines }
we compare DynaEval against three baselines: RANDOM, CoSim~\citep{xu-etal-2018-better} and S-DiCoh~\citep{mesgar-etal-2020-dialogue}. RANDOM baseline arbitrarily assigns a label to the input dialogue pairs. It suggests the peformance lower bound. CoSim is a common method for dialogue coherence assessment~\citep{xu-etal-2018-better,zhang2018reinforcing}. It obtains a dialogue-level score by averaging the cosine similarities between sentence embeddings of all adjacent utterance pairs within the dialogue. For fair comparison, we apply the same procedure described in Section~\ref{subsec:dur} to derive the sentence embedding of an utterance in CoSim. S-DiCoh~\citep{mesgar-etal-2020-dialogue} is a recent state-of-the-art dialogue coherence model. It models a dialogue with a neural network framework consisting of two bidrectional LSTM layers with attention mechanism at both the token and utterance level.

\textbf{Results and Analysis }
It can be observed in Table~\ref{tab:coh-results} that on all bench-marking dialogue datasets, DynaEval outperforms the baselines in both UR and SS category. Even though the dialogue datasets possess different characteristics as indicated in Section~\ref{subsec:dataset}, DynaEval exhbits robust performance across all the datasets. This confirms our hypothesis that DynaEval provides useful dialogue-level representation for distinguishing the original dialogues from the corresponding negative samples. Especially when compared to S-Dicoh, which models a dialogue sequentially with bidrectional LSTM and does not explicitly incoporate the speaker level interaction, the structured graph modeling of a dialogue in DynaEval is more effective for capturing both the interaction between the interlocutors and the contextual information within a dialogue. 

Based on the experimental results, it can be deduced that the discrimination task with UR strategy is more challenging compared to that with SS strategy. The accuracy scores achieved by S-DiCoh in the SS category is much higher than that in the UR category on both datasets. Similar observation can be made w.r.t CoSim and DynaEval on the ConvAI2 dataset. DynaEval performs remarkably in this task as it outperforms S-DiCoh by a significant margin of 13.97, 18.43 and 8.22 on Empathetic Dialogue, ConvAI2 and DailyDialog respectively. Given these observations, we further hypothesize that DynaEval model trained with UR strategy offers more useful dialogue representation to the dialogue evaluation task.
\begin{table*}	
\centering
\small
\resizebox{0.85\linewidth}{!}{
\begin{tabular}{@{}lcccccc|c@{}}
\toprule
\multicolumn{8}{c}{Dialogue-level Spearman Correlation} \\ \midrule
\textbf{Dialogue Aspects}  & \textbf{BERT-R} & \textbf{GPT-2} & \textbf{USR} & \textbf{S-DiCoh} & \textbf{FED} & \textbf{DynaEval} & \textbf{Human} \\ \midrule
Coherence & 0.229 & \textit{0.123} & 0.194 & \textit{0.038} & 0.251 & \textbf{0.423} & 0.809\\
Error Recovery & 0.242 & \textit{0.096}  & \textit{0.170} & \textit{-0.054} & \textit{0.165} & \textbf{0.311} & 0.840\\
Consistency & \textit{0.163}  & \textit{0.091}  &  \textit{0.169} & \textit{0.017} & \textit{0.116} & \textbf{0.352} & 0.562\\
Diversity & 0.196 & \textit{0.147}  & 0.242 & \textit{0.059} & \textbf{0.449} & 0.332 & 0.789\\
Topic Depth & 0.192 & \textit{0.097} & 0.341 & \textit{0.046} &  \textbf{0.522} & 0.439 & 0.833\\
Likability & 0.281  & 0.179 & 0.221 & \textit{-0.070} & 0.262 & \textbf{0.398} & 0.838\\
Understanding & 0.198 & \textit{0.070} & \textit{0.172} & \textit{-0.100} & 0.306 & \textbf{0.361} & 0.809\\
Flexibility & 0.253 & \textit{0.134} & 0.209 & \textit{0.044}   &  \textbf{0.408} & 0.389 & 0.816\\ 
Informativeness & 0.211 & \textit{0.116} & 0.288 & \textit{0.028} & 0.337  & \textbf{0.396} & 0.806\\
Inquisitiveness & 0.337 & \textit{0.071} & 0.188 & \textit{-0.054} & 0.298 & \textbf{0.388} & 0.769\\
\midrule
% Average & 0.124 & -0.133  & 0.061  & 0.122 & 0.105
% & \textbf{0.187} \\
Overall& 0.248 & \textit{0.123}  & 0.288 & \textit{-0.073} & 0.443 & \textbf{0.482} & 0.830\\ \midrule
\multicolumn{8}{c}{Turn-level Spearman Correlation} \\ \midrule
Interestingness & 0.235 &  -0.107  & \textit{0.085}  & \textit{0.031} & \textbf{0.431} & 0.289 & 0.819\\
Engagement &  0.206  & \textit{-0.086} & 0.107 & \textit{0.040}  & \textbf{0.318} & 0.255 & 0.798\\
Specificity & 0.327 & -0.112  & \textit{0.095} & \textit{0.062} & \textbf{0.326} & 0.272 & 0.790\\
Relevance & 0.151 & -0.105  & 0.183 & \textit{-0.051} & 0.152 & \textbf{0.265} & 0.753\\
Correctness & \textit{0.081} & \textit{0.041} & \textit{0.098} & \textit{-0.040} & 0.133 & \textbf{0.216} & 0.780\\
Semantically Appropriateness & \textit{0.044} & \textit{-0.084} & 0.201 & \textit{-0.069} & 0.177 & \textbf{0.233} & 0.682 \\
Understandable & \textit{0.051} & \textit{-0.071}  & 0.110 & \textit{-0.075} &0.111  & \textbf{0.185} & 0.522\\
Fluency & \textit{0.079} & -0.151 & 0.220& \textit{-0.007} & \textbf{0.224} & \textit{0.096} & 0.714\\ \midrule
% Average & 0.124 & -0.133  & 0.061  & 0.122 & 0.105
% & \textbf{0.187} \\
Overall & 0.195 & \textit{-0.095} & 0.137 & \textit{-0.022} & 0.209 & \textbf{0.264} & 0.820\\ \bottomrule
\end{tabular}}
% 	\end{subtable}
\caption{\label{tab:fed-correlation}Comparison of both dialogue and turn level Spearman correlations among state-of-the-art automatic metrics on the FED evaluation dataset. The results are reported for the 11 and 9 unique quality categories at turn and dialogue level respectively. Scores with p-values larger than 0.05 are italicized (indicating statistical insignificance). The best score for each category is highlighted in bold.}
\end{table*}

\subsection{Dialogue Evaluation Task}
\label{subsec:correlation-analysis}
To validate the above hypothesis, we assess the usefulness of DynaEval in both the dialogue-level and turn-level evaluation tasks. In both settings, Spearman correlations between the scores generated by DynaEval and the corresponding human evaluation scores are computed. The performance of DynaEval is compared against several recently proposed dialogue evaluators.

\textbf{Evaluation Dataset } FED~\citep{mehri-eskenazi-2020-unsupervised} is a bench-marking dataset useful for both dialogue-level and turn-level evaluation. It contains both human-human conversations and human-chatbot conversations, which are collected by the authors of the Meena chatbot~\citep{adiwardana2020towards} in an interactive setup. In total, 124 conversations are collected, out of which 40 come from interacting with the Meena Chatbot, 44 come from interacting with the Mitsuku Chatbot and 40 are drawn from human-human conversations. The average number of utterances per conversation is 13.72 and the average number of words per utterance is 9.23. Human quality annotations of these conversations are performed at both the dialogue and turn level. There are 9 quality aspects for turn-level annotations and 11 for dialog-level annotations outlined in the first column of Table~\ref{tab:fed-correlation}. FED includes 3348 turn-level and 1364 dialog-level annotations, for a total of 4712. The inter-annotator agreements for all the quality aspects, which indicate the metric performance upper bound, is shown in the last column of Table~\ref{tab:fed-correlation}. 

\textbf{Metrics to Compare} 
The recently proposed reference-free state-of-the-art dialogue metrics, including USR~\citep{mehri-eskenazi-2020-usr}, BERT-RUBER~\citep{ghazarian-etal-2019-better} (BERT-R), GPT-2 based coherence metric~\citep{pang-etal-2020-towards} (GPT-2) and FED~\citep{mehri-eskenazi-2020-unsupervised}\footnote{The correlation scores of FED is obtained from the original paper. For each evaluation category, the highest score is reported among the scores provided by all its variants.}, serve as the baseline dialogue evaluators. Since USR, BERT-R and GPT-2 are turn-level metrics, aggregation of all the turn-level scores in a dialogue is required for dialogue-level evaluation. The best correlation scores at dialogue level are reported in Table~\ref{tab:fed-correlation} among all the aggregation strategies for these three metrics. For completeness, we report their correlation scores w.r.t difference aggregation strategies in Appendix~\ref{subsec:turn-dl-corr}. Similar to DynaEval, S-Dicoh provides a unified score for each dialogue. Based on insights from Section~\ref{subsec:ddt}, the best performing model in the UR category is chosen to score the dialogues for both S-Dicoh and DynaEval.

\textbf{Dialogue-level Evaluation } 
DynaEval achieves the highest correlation scores in 8 out of 11 dialogue aspects, including the overall category. For the other three categories, DynaEval attains second highest correlation scores. We can see that DynaEval significantly outperforms S-DiCoh. These results showcase that structured graph modeling of a dialogue with explicit incorporation of speaker and utterance level dependencies provides meaningful dialogue-level representations. Such representations capture information of various dialogue attributes that are beneficial for the dialogue-level evaluation task.

Moreover, BERT-R, GPT-2 and USR are state-of-the-art turn-level evaluation metrics. They evaluate a dialogue based on aggregation of scores of all the context-response pairs within the dialogue. It can be observed that their correlation scores across individual dialogue aspects are not as high as those of DynaEval. This supports our hypothesis in Section~\ref{sec:introduction} that turn-level quality evaluation may be insufficient to assess the performance of open-domain dialogue systems. 

In addition, dialogue aspects, including coherence, likability, informativeness and Inquisitiveness, are highly dependent on the interaction of the interlocutors. Amongst all the dialogue aspects, DynaEval achieves significantly higher scores in these four categories. This attributes to its incorporation of the speaker level dependency. 

\textbf{Turn-level Evaluation} Furthermore, it can be observed that DynaEval achieves the highest correlation in 5 out of 9 categories including the overall category. This demonstrates that DynaEval is not only useful for holistic evaluation of a dialogue, but also useful for turn level evaluation. In this sense, DynaEval serves as a better proxy to the human evaluation process~\cite{li2019acute} whereby humans mainly evaluate the conversations in a holistic manner and laser-focus on the problematic turns.

Specifically, DynaEval performs well in turn-level aspects, such as relevance, semantic appropriateness and correctness. These aspects highly correlate to the dialogue-level attributes, such as coherence and understanding, suggesting that the evaluation of these turn-level attributes also benefit from the explicit modeling of the speaker and utterance level interaction in a unified framework.

\textbf{Error Analysis } An interesting finding is that DynaEval and FED actually complement each other at both dialogue and turn level. For example, at the dialogue level, FED performs well in diversity and topic depth, but struggles with coherence and consistency. DynaEval performs well in coherence and consistency, but its performance in diversity is much lower in comparison to FED. This may be because dialoGPT, the backbone of FED, was trained on a large amount of Reddit data, which contain diverse amount of topics and variation of expressions while DynaEval is trained on a single dialogue domian. Moreover, dialoGPT does not explicitly model such speaker-level interaction, but DynaEval does. Hence, DynaEval is more useful for evaluating coherence and consistency aspects of a dialogue. One way to improve DynaEval for evaluating topic depth and diversity is to pre-train on a large amount of dialogue data with a variety of topics and then fine-tune it on the target domain.

Another observation is that DynaEval performs significantly poorer for the fluency aspect at turn-level than for other turn-level aspects. Additionally, GPT-2, USR and FED, which leverage pretrained language model, perform significantly better than DynaEval in this category. This may be because DynaEval directly models a dialogue at the utterance level instead of at the token level, while the other metrics consider the language modeling objective, which focuses more on the token-level dependencies rendering them effective for evaluating the naturalness of a response. A remedy to this problematic aspect of DynaEval is to introduce perturbation strategies targeting the token level, such as word drop, word shuffling and word replacement~\citep{sinha-etal-2020-learning,park-etal-2021-generating}. Such strategies provide negative samples mimicking the non-sensical or non-grammatical responses produced by certain seq2seq generative models. Another simple solution is to combine DynaEval with turn-level metrics specifically designed for evaluating naturalness of dialogue responses.     

Besides the fluency aspect, DynaEval's performance in interestingness, engagement and specificity at the turn level is not as pronounced as that of FED. This may be because purely modeling the dialogue itself is not enough for all the aspects. The model may need to incorporate external knowledge concerning a diverse range of topics to be able to reflect these attributes. The same conclusion can also be drawn from DynaEval's relatively weaker performance in the diversity category at the dialogue level.  

Lastly, DynaEval primarily targets open-domain dialogues where there is no clear or predefined task to perform. When evaluating task-oriented dialogues, task completion will take a more central role. Meta-information such as intents and request types are important to determine task completion and therefore, the evaluation framework will require further adaptation accounting for these information when evaluating task-oriented dialogues.  

\section{Conclusion \& Future Work}
\label{sec:conclusion}

DynaEval serves as a unified framework for both turn and dialogue level evaluation in open-domain dialogue. It provides meaningful representations that incorporate information reflecting various important dialogue attributes. Its explicit modeling of speaker and utterance level interaction leveraging GCN has been proven beneficial for the evaluation task. Lastly, the error analysis in Section~\ref{subsec:correlation-analysis} sheds light on how DynaEval can be further improved. DynaEval can also be combined with the specialized turn-level metrics, such as those targeting fluency and engagement, to fully approximate the interactive human evaluation process.

\section*{Acknowledgement}
\label{sec:ack}
We would like to thank all the anonymous reviewers for their constructive comments. This work is supported by Human-Robot Interaction Phase 1 (Grant No. 19225 00054), National Research Foundation (NRF) Singapore under the National Robotics Programme; Human Robot Collaborative AI for AME (Grant No. A18A2b0046), NRF Singapore; Robert Bosch (SEA) Pte Ltd under EDB’s Industrial Postgraduate Programme – II (EDB-IPP), project title: Applied Natural Language Processing; and by the Spanish projects: AMIC (MINECO, TIN2017-85854-C4-4-R) and CAVIAR (MINECO, TEC2017-84593-C2-1-R) projects partially funded by the European Union.

\section*{Ethical Considerations \& Broader Impact}
\label{sec:ec}
This study conforms to the prevailing ethical guidelines. All datasets used are in the public domain. In addition, we have identified a way that DynaEval can help address the ethical concerns. By explicitly training the framework to discriminate safe dialogues from unsafe ones, it can help detect dialogues containing inappropriate sentences, such as those regarding injustice and discrimination. Such application may be useful in many real-life scenarios where the behaviors of chatbots need to be properly monitored to avoid insensitive and irresponsible comments from the chatbots.

\newpage
\appendix

\section{Additional Experimental Results}

\subsection{Utterance-level Pooling Techniques} 
To derive the dialogue-level representation, we have adopted the mean pooling method in DynaEval. In this section, we examine the effects of different pooling methods in the dialogue-level discrimination task. Specifically, we compare the performance of mean pooling against max pooling and the concatenation of sentence vectors derived with both mean and max pooling. The performance comparison is presented in Table~\ref{tab:pool-coh-results}. It can be observed that the performance difference across various pooling strategies is not statistically significant.  

\begin{table}[!ht]
\centering
\begin{tabular}{ccc}
\toprule
Strategy & UR & SS \\ 
\midrule
Mean & 94.30 ± 0.07 &  90.37 ± 0.37  \\
Max  &  94.17 ± 0.16 &  90.75 ± 0.24   \\
Mean+Max & 94.19 ± 0.04 & 90.64 ± 0.06   \\ 
\bottomrule
\end{tabular}
\caption{The accuracy scores (\%) of DynaEval on the test set of Empathetic Dialogue with different utterance-level pooling techniques. The average and standard deviation are reported in the table.}
\label{tab:pool-coh-results}
\end{table} 

\subsection{Dialogue-level Correlation Analysis of Turn-level Metrics} 
\label{subsec:turn-dl-corr}
For each turn-level metric, we have applied four simple aggregation strategies to derive dialogue level scores from their respective constituent turn level scores: (1) Mean, (2) Sum, (3) Max and (4) Multiplication. The dialogue level correlation coefficients of USR, BERT-RUBER and GPT-2 based coherence metric are reported in Table~\ref{tab:usr-turn-corr}, Table~\ref{tab:bertr-turn-corr} and Table~\ref{tab:gpt2-turn-corr} correspondingly. Note that for turn-level metrics leveraging the language model objective, we don't consider token-level aggregation variants. Instead, we follow the same formulations in the original papers. For example, the GPT-2 based coherence metric~\citep{pang-etal-2020-towards} computes a turn-level score based on averaging the token-wise conditional log probabilities in the corresponding response. 

It can be observed that all three metrics don't perform well at dialogue level evaluation. This further validates our statement in Section~\ref{sec:introduction} that turn-level quality evaluation may be insufficient to assess the performance of open-domain dialogue systems as they don't specifically model the interaction over an entire dialogue.

\section{Reproducibility}

\begin{table}[!t]
\centering
\resizebox{\linewidth}{!}{
    \begin{tabular}{c|c|c|c|c} \midrule
        \textbf{Quality} & \textbf{Mean} & \textbf{Sum} & \textbf{Max} & \textbf{Prod} \\ \midrule
\multicolumn{5}{c}{USR} \\ \midrule
Coherence & \textbf{0.194} & \textit{0.111} & \textit{0.021} & \textit{0.158}\\
Error Recovery & \textbf{\textit{0.170}} & \textit{0.083}  & \textit{0.075} & \textit{0.130} \\
Consistency & \textit{0.150} & \textbf{\textit{0.169}} & \textit{0.038} & \textit{0.099}\\
Diversity & \textbf{0.242} & \textit{0.167} & 0.235 & 0.193\\
Topic Depth & \textbf{0.341} & \textit{0.145} & 0.255 & 0.295\\
Likability & \textbf{0.221} & 0.193 & \textit{0.109} & \textit{0.126}\\
Understanding & \textbf{\textit{0.172}} & \textit{0.112} & \textit{0.004} & \textit{0.124}\\
Flexibility & \textbf{0.209} & \textit{0.151} & \textit{0.164} & \textit{0.129}\\
Informativeness & \textbf{0.288} & \textit{0.157} & \textit{0.171} & 0.237\\
Inquisitiveness & 0.148 & \textit{0.099} & \textbf{0.188} & \textit{0.128}\\ \midrule
Overall & \textbf{0.288} & \textit{0.166} & \textit{0.094} & 0.212\\\bottomrule
\end{tabular}
}
\caption{\label{tab:usr-turn-corr}Dialogue level Spearman correlation coefficients of USR w.r.t different turn-level aggregation strategies on the FED dataset. Scores with p-values larger than 0.05 are italicized (indicating statistical insignificance). The best score for each category is highlighted in bold.}
\end{table}

\begin{table}[!t]
    \centering
\centering
\resizebox{\linewidth}{!}{
    \begin{tabular}{c|c|c|c|c} \midrule
        \textbf{Quality} & \textbf{Mean} & \textbf{Sum} & \textbf{Max} & \textbf{Prod} \\ \midrule
\multicolumn{5}{c}{BERT-R} \\ \midrule
Coherence & 0.222 & 0.221 & \textbf{0.229} & \textit{0.041} \\
Error Recovery & 0.231 & \textbf{0.242} & 0.228 & \textit{0.005} \\
Consistency & \textit{0.141} & \textbf{\textit{0.163}} & \textit{0.148} & \textit{-0.030} \\
Diversity & 0.180 & \textbf{0.196} & \textit{0.164} & \textit{-0.051} \\
Topic Depth & 0.181 & \textbf{0.192} & \textit{0.163} & \textit{0.008}\\
Likability & 0.256 & \textbf{0.281} & 0.249 & \textit{-0.037} \\
Understanding & 0.189 & \textbf{0.198} & 0.189 & \textit{-0.023}\\
Flexibility & 0.228 & \textbf{0.253} & 0.232 & \textit{-0.036} \\
Informativeness & 0.194 & \textbf{0.211} & 0.186 & \textit{-0.023} \\
Inquisitiveness & 0.326 & \textbf{0.337} & 0.331 & \textit{0.056} \\ \midrule
Overall & 0.231 & \textbf{0.248} & 0.224 & \textit{-0.021} \\\bottomrule
\end{tabular}
}
\caption{\label{tab:bertr-turn-corr}Dialogue level Spearman correlation coefficients of BERT-RUBER w.r.t different turn-level aggregation strategies on the FED dataset.}
\end{table}

\begin{table}[!t]
\centering
\resizebox{\linewidth}{!}{
    \begin{tabular}{c|c|c|c|c} \midrule
        \textbf{Quality} & \textbf{Mean} & \textbf{Sum} & \textbf{Max} & \textbf{Prod} \\ \midrule
\multicolumn{5}{c}{GPT-2} \\ \midrule
Coherence & \textit{-0.002} & \textbf{\textit{0.123}} & \textit{-0.086} & \textit{-0.120} \\
Error Recovery & \textit{0.034} & \textbf{\textit{0.096}} & \textit{-0.057} & \textit{-0.091} \\
Consistency & \textit{-0.025} & \textbf{\textit{0.091}} & \textit{-0.048} & \textit{-0.088}\\
Diversity & \textit{0.092} & \textit{\textbf{0.147}} & \textit{-0.033} & \textit{-0.145}\\
Topic Depth & \textit{0.054} & \textbf{\textit{0.097}} & \textit{-0.036} & \textit{-0.094} \\
Likability & \textit{0.072} & \textbf{0.179} & \textit{-0.047} & \textit{-0.175} \\
Understanding & \textit{-0.027} & \textbf{\textit{0.070}} & \textit{-0.062} & \textit{-0.066}\\
Flexibility & \textit{0.056} & \textit{\textbf{0.134}} & \textit{-0.032} & \textit{-0.131}\\
Informativeness & \textit{0.025} & \textbf{\textit{0.116}} & \textit{-0.100} & \textit{-0.112} \\
Inquisitiveness & \textit{-0.008} & \textit{\textbf{0.071}} & \textit{-0.071} & \textit{-0.070} \\ \midrule
Overall & \textit{-0.002} & \textbf{\textit{0.123}} & \textit{-0.086} & \textit{-0.120} \\\bottomrule
\end{tabular}
}
\caption{\label{tab:gpt2-turn-corr}Dialogue level Spearman correlation coefficients of GPT-2 based coherence metric w.r.t different turn-level aggregation strategies on the FED dataset.}
\end{table}

\subsection{Training Setup \& Hyperparameters}
For all the experiments involving training, we run the experiments five times with different random seeds for model weights initialization to reduce the risk of randomness. The experiments are performed on a single Tesla V100 32GB GPU with a batch size of 512. The model is trained for 20 epochs and its parameters are optimized using the Adam optimizer. The average run time for each epoch is around 8 hours and 15 minutes. The initial learning rate is set to 0.002 and decays by a factor of 0.5 per epoch. A dropout of 0.5 is also applied. 

For Empathetic Dialogue and DailyDialog, the context window length, $M$ is set to 4, because these two datasets contain relatively short conversations (4.31 and 7.90 average number of utterances per dialogue respectively). A context window size of 4 ensures each utterance is connected to all the remaining utterances in most of the dialogues. The utterances may provide important contextual information to each other within a dialogue. For ConvAI2, $M$ is set to 2 to avoid introducing too much irrelavant context information. This is because most of the conversations in ConvAI2 are about two people getting to know each other and there are frequent topic changes in the conversations. $M$ serves as an important hyperparameter to control the influence of an utterance on the rest in a dialogue.

For training DynaEval, we have filtered out dialogues of which the number of utterances is less than 4 or more than 30. We hypothesize that dialogues with less than 4 utterances containing little information for modeling speaker and utterance level interaction. Moreover, there are very few dialogues with more than 30 utterances in both datasets. Including them leads to large graphs and unnecessary paddings, which slow down the training process.

\end{document}